\documentclass[10pt]{article}
\newif\ifblindreview

\usepackage[letterpaper]{geometry}
\usepackage{amta2024}
\usepackage{times}
\usepackage{url}
\usepackage{latexsym}
\usepackage{natbib}
\usepackage{layout}
\usepackage{multicol}
\usepackage{booktabs,array}
\usepackage{float}
\usepackage[hidelinks]{hyperref}
\usepackage[all]{hypcap}
\usepackage{graphicx}
\usepackage{subfig}
\usepackage{inconsolata}
\usepackage{xcolor}
\usepackage{amsmath}
\usepackage{tabularx}

\hypersetup{
    colorlinks,
    linkcolor={red!50!black},
    citecolor={blue!50!black},
    urlcolor={blue!80!black}
}

\ifblindreview
  \newcommand{\authorinfo}{\author{}} 
\else
  \newcommand{\authorinfo}{
    \author{\name{\bf Inacio Vieira\footnote{Equal contribution}} \hfill  \addr{inacio@gmail.com}\\
            \addr{\small Department of Computing, Dublin City University, Dublin, Ireland \& Alpha CRC, Cambridge, UK}
    \AND
            \name{\bf Will Allred$^*$} \hfill \addr{william.allred2@mail.dcu.ie}\\
            \addr{\small Department of Computing, Dublin City University, Dublin, Ireland}
    \AND
           \name{\bf Séamus Lankford} \hfill \addr{seamus.lankford@adaptcentre.ie}\\
            \addr{\small Department of Computer Science, Munster Technological University, Cork, Ireland}
    \AND            
           \name{\bf Sheila Castilho} \hfill \addr{sheila.castilho@adaptcentre.ie}\\
            \addr{\small SALIS/ADAPT Centre, Dublin City University, Dublin, Ireland}
    \AND 
           \name{\bf Andy Way} \hfill \addr{andy.way@adaptcentre.ie}\\           
            \addr{\small ADAPT Centre, School of Computing, Dublin City University, Dublin, Ireland}
    }
  }
\fi

\begin{document}

\amtaHeader{x}{x}{xxx-xxx}{2015}{45-character paper description goes here}{Author(s) initials and last name go here}
\title{\bf How Much Data is Enough Data? Fine-Tuning Large Language Models for In-House Translation: Performance Evaluation Across Multiple Dataset Sizes}
\authorinfo

\maketitle

\pagestyle{empty}

\begin{abstract}
\vspace{7pt}

Decoder-only LLMs have shown impressive performance in MT due to their ability to learn from extensive datasets and generate high-quality translations. However, LLMs often struggle with the nuances and style required for organisation-specific translation. In this study, we explore the effectiveness of fine-tuning Large Language Models (LLMs), particularly Llama 3 8B Instruct, leveraging translation memories (TMs), as a valuable resource to enhance accuracy and efficiency.

We investigate the impact of fine-tuning the Llama 3 model using TMs from an organisation in the software sector. Our experiments cover five translation directions across languages of varying resource levels (English to Brazilian Portuguese, Czech, German, Finnish, and Korean). We analyse diverse sizes of training datasets (1k to 207k segments) to evaluate their influence on translation quality. We fine-tune separate models for each training set and evaluate their performance based on automatic metrics, BLEU, chrF++, TER, and COMET.

Our findings reveal improvement in translation performance with larger datasets across all metrics. On average, BLEU and COMET scores increase by 13 and 25 points, respectively, on the largest training set against the baseline model. Notably, there is a performance deterioration in comparison with the baseline model when fine-tuning on only 1k and 2k examples; however, we observe a substantial improvement as the training dataset size increases. The study highlights the potential of integrating TMs with LLMs to create bespoke translation models tailored to the specific needs of businesses, thus enhancing translation quality and reducing turn-around times. This approach offers a valuable insight for organisations seeking to leverage TMs and LLMs for optimal translation outcomes, especially in narrower domains.

\end{abstract}

\begin{multicols}{2}
\section{Introduction}

\begin{table*}[ht!]
\centering
\begin{tabular*}{\textwidth}{@{\extracolsep{\fill}} lr}
\toprule
\textbf{Datasets} & \textbf{Segments} \\
\midrule
Aligned Training & 1000, 2000, 5000, 10000, 14688 \\
Dev & 1837 \\
Test & 1353 \\
\bottomrule
\end{tabular*}
\caption{Segment counts for the various aligned training dataset sizes, the development set, and the test set.}
\label{tab:data-counts-aligned}
\end{table*}

In recent years, decoder-only large language models (LLMs) have revolutionised the machine translation (MT) field due to their ability to learn from vast amounts of data and generate high-quality translations \citep{alves-etal-2023-steering, moslem_adaptive_2023, mu-etal-2023-augmenting, robinson-etal-2023-chatgpt, zhu_multilingual_2023, lyu_paradigm_2024}. LLMs, such as Llama 3 8B Instruct,\def\thefootnote{*}\footnotetext{These authors contributed equally to this work}\def\thefootnote{\arabic{footnote}}\footnote{\url{https://huggingface.co/meta-llama/Meta-Llama-3-8B-Instruct}} have shown impressive capabilities in adapting to translation tasks, generating human-like accurate output, making them invaluable tools for the sector \citep{li_eliciting_2023, moslem_fine-tuning_2023, lyu_paradigm_2024}. However, out-of-the-box LLMs do not always capture all the nuances, appropriate tone, and terminology required for specialised or organisation-specific translations \citep{moslem_domain-specific_2022, alves_steering_2023, zheng_fine-tuning_2024}. This is where translation memories (TMs) offer a potential solution.

A TM is a database that stores previously human-translated segments and their respective translations. They are particularly useful to language service providers (LSPs) as they deal with repetitive content and organisation-specific style and terminology, enhancing the efficiency and accuracy of translations \citep{bloodgood_translation_2014,bulte_neural_2019,moslem_adaptive_2023}. Therefore, the integration of TMs and LLMs can create models that better understand organisational requirements and lead to higher quality outputs and reduced turnaround times. However, this approach depends on several factors, like the amount, quality and specificity of the TMs used as training data for fine-tuning.

Previous work explored fine-tuning of models with TM for translation for specific domains and the benefit that offers to performance \citep{Haque2020-Terminology, moslem_domain-specific_2022}. Accordingly, TM provides much value because of its high quality and domain relevance \citep{bulte_neural_2019, xu_boosting_2020,cai_neural_2021, knowles_translation_2022}. This research highlights the gains available when leveraging existing TMs during the fine-tuning process of LLMs.

In this paper, we investigate a real-life scenario where we fine-tune Llama 3 8B Instruct \citep{llama_team_llama_2024} using TMs from a specific organisation. Additionally, since increasing the fine-tuning data requires dedicating more resources and time, we explore different dataset sizes to evaluate their impact on translation quality and identify the most efficient return on investment. We conduct experiments in five translation directions (from English) on languages of varying resource level (Brazilian Portuguese (PT-BR), Czech (CS), German (DE), Finnish (FI), and Korean (KO)). This approach can lead to bespoke translation models that cater to the unique needs of different companies when compared to generic LLMs.

\begin{table*}[!ht]
\centering
\begin{tabular*}{\textwidth}{@{\extracolsep{\fill}} lrrr}
\toprule
\textbf{Lang} & \textbf{Full Training Data} & \textbf{Dev Data} & \textbf{Total Segments} \\
\midrule
Brazilian Portuguese (PT-BR)   & 217,555 & 54,389  & 271,944 \\ 
Czech (CS)           & 107,555 & 26,889  & 134,804 \\ 
German (DE)          & 223,894 & 55,973  & 279,867 \\ 
Finnish (FI)         & 207,218 & 51,805  & 259,023 \\ 
Korean (KO)          & 162,360 & 40,590  & 202,950 \\ 
\bottomrule
\end{tabular*}
\caption{Segment counts for the full datasets used during training.}
\label{tab:data-counts-full}
\end{table*}

\section{Methodology}
\subsection{Data}

The raw dataset consists of TMs from an anonymous organisation that operates in the software sector. The three datasets employed cover knowledge base, mobile user interface, and mobile reference materials. 

The five target languages dataset (PT-BR, CS, DE, FI, and KO) are filtered to remove duplicates, source-copies, and segments over 150 words to ensure none would go over the maximum length set during training. All HTML tags are removed, and double spaces are converted to single spaces. Any rows containing only dates, version numbers, or any programming language are also removed. Rows are then randomly shuffled to mitigate any temporal bias that could arise from the chronological order of the data, ensure the model does not memorise sequences, and prevent the evaluation set from being biased towards a particular section of the data.

The dataset is then transformed into an inter-lingual aligned dataset for all five target languages where any rows with missing translations for any target languages, are dropped. This results in a dataset where all source segments have translations available in all five target languages. The dataset is then split into training, development, and test sets, as shown in \mbox{Table \ref{tab:data-counts-aligned}}. 

Further filtering is applied to the test set removing segments that had over 75\% similarity with any segments in the training dataset to ensure robust testing and minimal memorisation. We measure similarity as a combination of the Levenshtein distance \citep{levenshtein_binary_1965} and a 5-gram-based similarity \citep{lopez-gazpio_word_2019}. This reduced the size of the test split from 1837 to 1353. The test split with under 75\% similarity was used for all experiments.

In the interest of using all the data available, we also compile all segments in a given language into a dataset for each target language. This includes any segment that would not fit the inter-lingual alignment criteria applied above. This will now be referred to as the ‘full dataset’. These larger training sets allow us to train beyond the 14.7k aligned segments and make use of the total volume of available segments in order to explore what impact that would have on results. The full training sets range from 107k (CS) to 223k (DE) examples, as shown in Table \ref{tab:data-counts-full}.

\subsection{Model}

We use the Llama 3 8B Instruct model and its associated okenizer \citep{llama_team_llama_2024}. The decision between the Instruct and the base model is based on an extensive MT evaluation of Llama 3 models \citep{Wu2024-Document-MT} using the Flores-200\footnote{\url{github.com/facebookresearch/flores/blob/main/flores200/README.md}} dataset \citep{guzman-etal-2019-flores, NLLB2022}. Even though \cite{wu_evaluating_2024} dealt with the opposite language direction (X to English), we consider the close results between Instruct and the base model involving the five languages included in this paper to be a good indicator of proximity in performance between the models.
Our baseline consists of the test set metric results obtained from the out-of-the-box Llama 3 8B Instruct model.
We use QLoRA \citep{hu_lora_2021,dettmers_qlora_2023} for efficient fine-tuning with 4-bit quantisation using Hugging Face Transformers. We perform fine-tuning on a high performance cluster with four A100-SXM4-80GB GPUs. From Hugging Face, we leverage the  Supervised Fine-Tuning Training (SFTTrainer),\footnote{\url{https://huggingface.co/docs/trl/en/sft_trainer}} which is a wrapper of the Trainer class\footnote{\url{https://huggingface.co/docs/transformers/en/main_classes/trainer}} optimized for fine-tuning language models like Llama. On the largest dataset size, fine-tuning takes approximately 2.3 hours (Appendix \ref{sec:appendix-a}).

\begin{table*}[!ht]
\centering
\begin{tabular*}{\textwidth}{@{\extracolsep{\fill}} ll}
\toprule
\multicolumn{2}{l}{\textbf{BitsAndBytes Quantisation Configuration}} \\
\midrule
load\_in\_4bit                        & True \\
bnb\_4bit\_quant\_type                 & ``nf4" \\
bnb\_4bit\_use\_double\_quant           & True \\
bnb\_4bit\_compute\_dtype              & torch.bfloat16 \\
\midrule
\multicolumn{2}{l}{\textbf{PEFT LoRA Configuration}} \\
\midrule
low-rank matrix dimension (r)       & 64 \\
scaling factor (lora\_alpha)         & 16 \\
dropout probability (lora\_dropout)  & 0.1 \\
training of bias parameters (bias)  & ``none" \\
\midrule
\multicolumn{2}{l}{\textbf{Training Arguments}} \\
\midrule
batch size for training and evaluation & 32 examples \\
learning rate                        & 2e-3 \\
lr\_scheduler\_type                    & ``constant" \\
bf16                                 & True \\
\bottomrule
\end{tabular*}
\caption{Fine-tuning hyperparameters.}
\label{tab:hyperparameters}
\end{table*}

\subsection{Inference}

\subsubsection{Prompting}

At inference time, we use many of the recommended parameters from previous work \citep{moslem_fine-tuning_2023} and model documentation to produce translation outputs from the baseline model and the fine-tuned versions (cf. Appendix \ref{sec:appendix-c}). Meta’s Llama 3 documentation\footnote{\url{https://llama.meta.com/docs/model-cards-and-prompt-formats/meta-llama-3/}} provides a recommended prompt format and instructions to implement special tokens during inference and training \citep{llama_team_llama_2024}.

The prompt and the source segment were passed to the model for inference to obtain each translation. This constitutes zero-shot as it did not include examples in the prompt \citep{Zhang2023-MT-Efficient-Finetuning}. A JSON scheme (\textit{\{``translation'': ``string''\}}) was also added to the prompt in order to obtain a structured output \citep{Wu2024-Document-MT}. During training, the same format was applied with the addition of the specific EOS token (\mbox{\textit{$<|$end\_of\_text$|>$}}) as recommended by Meta’s documentation (cf. Appendix \ref{sec:appendix-b}).

\subsubsection{Translation}

In order to obtain higher efficiency, both baseline and fine-tuned models are converted to the CTranslate2\footnote{\url{https://github.com/OpenNMT/CTranslate2}} \citep{Klein2020-Efficient} format (with 8-bit quantisation) and provided with parameters for inference (cf. Appendix \ref{sec:appendix-c}).

\subsubsection{Stopping Criteria and Post-processing}

In early experiments, we observe frequent instances of overgeneration; an issue recently explored further by \cite{zheng_fine-tuning_2024}. By using "\}assistant" as a \textit{stop\_token} in our \textit{stopping\_criteria}, we find much less post-processing is required in order to obtain the pure translation.

Our post-processing consists of extracting the translation by removing the `\{\textit{``translation'': ``}\,' prefix and the trailing `\,\textit{''\}}\,'. The newline characters are replaced by spaces. On some occasions, especially in the models produced by the smaller training datasets (1k and 2k examples), further cleaning is required as the model inadvertently overgenerated some HTML tags like ‘$<$br$>$’ and ‘$<$p$>$’. This is important to properly assess the translation quality.

\subsection{Evaluation}
 
To evaluate the performance of our models, we report BLEU \citep{Papineni2002-BLEU}, chrF++  \citep{Popovic2017-chrF++}, TER \citep{Snover2006-TER} via sacreBLEU,\footnote{\url{https://github.com/mjpost/sacrebleu}} and COMET\footnote{wmt20-comet-da, \url{https://github.com/Unbabel/COMET}} \citep{Rei2020-COMET}. We use multiple metrics to make our experiments more comparable to a wider variety of work and to provide insight into certain aspects of performance. 

It is important to note that the experiment aims to show the training efficiency of the PEFT fine-tuning method and its ability to approximate the model’s translating capabilities to the training material. Therefore, we pay special attention to the automatic metrics measuring \textit{n}-gram differences and edits (BLEU, chrF++, TER) whilst still considering the quality estimation aspect of COMET as a means of comparing inter-source languages and other similar research. Our results are compared to those obtained from the baseline model, an out-of-the-box Llama 3 8B Instruct model, and to GPT 3.5. We also ask five professional translators to post-edit 100 translations from the best-performing model into their language pair. They also answer a questionnaire about the quality of the automatically translated segments. The questionnaire asks for comments on the quality of the translations.

\section{Results and Discussion}

The results in Table~\ref{tab:results} show an increase in performance across all the languages for all datasets with more than 5k segments compared to the baseline. The fully aligned 14.7k dataset sees a BLEU score increase of 4.8 points or relative increase of 17.42\% on average over the baseline, over all target languages, while chrF++ and COMET increases 7.1 and 16.9, respectively. Similarly, TER decreases by 9 points. The 100k+ datasets also demonstrate consistent performance gains with an average increase of 13.7 BLEU, 12.7 chrF++, and 25 COMET, while TER decreases to 15.5.

To provide a point of comparison, we evaluate the performance of GPT-3.5\footnote{\url{https://chat.openai.com/}} on our test set. While GPT-3.5 outperforms our highest-performing model in BLEU and chrF++ for DE and FI, the 100k+ datasets often surpass GPT-3.5 in other languages and metrics. This demonstrates the effectiveness of creating bespoke models through fine-tuning mid-sized LLMs when leveraging domain-specific data. Targeted fine-tuning can yield competitive or superior results compared to larger, general-purpose models like GPT-3.5.

\begin{table*}[!ht]  
\centering
\begin{small}
\begin{tabular}{l*{5}{>{\raggedleft\arraybackslash}p{5em}}}
\textbf{Lang} & \textbf{Data Size} & \textbf{BLEU $\uparrow$} & \textbf{chrF++ $\uparrow$} & \textbf{TER $\downarrow$} & \textbf{COMET $\uparrow$} \\
\toprule
& GPT 3.5 & 56.50 & 76.33 & 32.03 & 86.02 \\
& Baseline & 48.25 & 69.21 & 39.36 & 77.28 \\
& 1k & 48.00 & 69.34 & 40.11 & 78.28 \\
\textbf{PT-BR} & 2k & \underline{46.04} & \underline{67.93} & \underline{44.09} & \underline{75.70} \\
& 5k & 49.73 & 69.92 & 38.03 & 80.80 \\
& 10k & 50.90 & 70.92 & 35.96 & 86.15 \\
& 14.7k & 53.42 & 73.07 & 32.92 & 89.18 \\
& 100k+ & \textbf{62.45} & \textbf{78.57} & \textbf{26.20} & \textbf{95.98} \\
\midrule
& GPT 3.5 & 31.78 & 55.02 & 58.17 & 72.99 \\
& Baseline & 26.25 & 49.97 & 63.27 & 62.43 \\
& 1k & 26.53 & 50.15 & 64.97 & 64.20 \\
\textbf{CS} & 2k &  \underline{25.23} & \underline{48.35} & \underline{68.76} & \underline{58.28} \\
& 5k & 27.57 & 51.35 & 62.84 & 66.85 \\
& 10k & 27.96 & 52.40 & 63.26 & 66.62 \\
& 14.7k & 31.57 & 54.75 & 60.07 & 73.73 \\
& 100k+ & \textbf{39.72} & \textbf{61.45} & \textbf{52.00} & \textbf{84.22} \\
\midrule
& GPT 3.5 & \textbf{42.41} & \textbf{65.88} & 50.07 & 65.31 \\
& Baseline & 34.32 & 59.16 & 57.60 & 58.36 \\
& 1k & 34.58 & 59.07 & 58.42 & 60.86 \\
\textbf{DE} & 2k & \underline{32.45} & \underline{57.08} & \underline{62.93} & \underline{53.87} \\
& 5k & 35.31 & 59.37 & 56.19 & 63.66 \\
& 10k & 37.23 & 60.58 & 53.59 & 66.82 \\
& 14.7k & 37.88 & 61.08 & 52.71 & 68.50 \\
& 100k+ & 42.27 & 65.15 & \textbf{48.59} & \textbf{73.01} \\
\midrule
& GPT 3.5 & \textbf{33.80} & \textbf{59.18} & \textbf{58.29} & 83.84 \\
& Baseline & 23.97 & 49.70 & 70.36 & 62.64 \\
& 1k & 24.14 & 49.48 & 71.18 & 65.22 \\
\textbf{FI} & 2k & \underline{19.07} & \underline{46.97} & \underline{83.00} & \underline{58.34} \\
& 5k & 22.05 & 47.32 & 75.12 & 60.54 \\
& 10k & 25.88 & 50.71 & 65.99 & 74.91 \\
& 14.7k & 26.48 & 51.32 & 64.91 & 73.66 \\
& 100k+ & 31.71 & 57.13 & 59.72 & \textbf{84.71} \\
\midrule
& GPT 3.5 & 33.07 & 49.72 & 60.60 & 63.28 \\
& Baseline & 20.81 & \underline{35.37} & 77.95 & 36.45 \\
& 1k & 20.12 & 42.16 & \underline{83.37} & 35.24 \\
\textbf{KO} & 2k & \underline{19.25} & 41.13 & 82.48 & \underline{26.03} \\
& 5k & 28.60 & 46.84 & 65.42 & 54.17 \\
& 10k & 31.36 & 52.62 & 60.86 & 70.56 \\
& 14.7k & 28.15 & 58.88 & 53.11 & 76.65 \\
& 100k+ & \textbf{45.80} & \textbf{64.81} & \textbf{44.73} & \textbf{84.30} \\
\bottomrule
\end{tabular}
\end{small}
\caption{Evaluation results of fine-tuning Llama 3 8B on datasets of various sizes. \textbf{Bold} text indicates the best score. The models trained on the largest dataset (100k+) perform the best. The scores deteriorate from the baseline for 1k and 2k but recover and increase from 5k onward. \underline{Underlined} text indicates the worst scores.}
\label{tab:results}
\end{table*}

\subsection{Small Dataset Deterioration}

Regarding translation quality across different training data sizes, we note a deterioration in quality for models trained on the smaller datasets (1k and 2k) in relation to the baseline. Despite a smooth reduction in both training and evaluation loss during training across all sizes, these smaller datasets still lead to poorer performance on all metrics. This can be due to the fact that the 1k and 2k datasets are insufficient to offer the models a wide enough variety of examples, leading to overfitting where the model performs well on training but poorly on the unseen test data \citep{barone_regularization_2017, atrio_interaction_2022, garcia_unreasonable_2023, ramirez_atrio_regularization_2023}.

It is possible that the lack of diversity in the smaller models fails to capture the range of linguistic and translation nuances present in the test data which hinders the model’s ability to generalise beyond the specific examples seen during training. Furthermore, the smaller datasets may make the models more susceptible to noise, such as translation errors or inconsistencies, leading to the learning of incorrect patterns and degrading performance on the test data, affecting the automatic metrics results, while the loss continues to drop due to fitting noisy data \citep{barone_regularization_2017, atrio_interaction_2022, ramirez_atrio_regularization_2023}.

Another possible explanation for the deterioration is a decrease in training data quality in the 1k and 2k dataset sizes. To examine this, we use COMET-Kiwi \citep{rei_scaling_2023}, a popular quality estimation metric, to evaluate the quality of the training data. The scores are consistent for each language with variations within a narrow range of 1-2 points (cf. Appendix \ref{sec:appendix-d}). For example, FI has the highest variation with a maximum score of 79.58 (1k and 14.7K) and a minimum score of 78.12 (5k), resulting in a range of only 1.46 points. The minimal variation in score indicates consistent data quality across all dataset sizes for each language. Therefore, the deterioration in performance is unlikely to be due to a decrease in data quality for the 1k and 2k training data sizes.

Hyperparameter fine-tuning could be employed to mitigate this early deterioration in situations where only small datasets are available. This may include dropout or other regularisation techniques to prevent overfitting on small training sets. Adjustment of the learning rate, batch sizes and QLoRA hyperparameters should also be explored to deal with this specific case of deterioration \citep{barone_regularization_2017, atrio_interaction_2022, dettmers_qlora_2023, ramirez_atrio_regularization_2023}.

Overall, a different approach is required in order to obtain gains when the training data is scarce. Our experiments suggest the need for at least 5k examples to achieve an improvement in metrics under the hyper-specific domain and circumstances we explore.

The issues above seem to be mitigated on the larger sets whilst maintaining the same hyperparameters as previously reported (cf. Table~\ref{tab:hyperparameters}). We observe performance recovery on 5k examples, overtaking the baseline model, then consistently improving over all metrics as dataset size increases, and achieving increasingly impressive results across all metrics when training on anything above the 10k sets and excelling on the 100k+ sets. 

\subsection{Resource Level}

It is interesting to note that the performance for KO has improved after the 14.7k fine-tuning and becomes comparable to or better than the performance of the other language directions, despite the lower initial baseline score across all metrics. For instance, the COMET score for the KO baseline is 36.5 while the average for all other languages is 57.7. We find that the lower resource languages (KO being the lowest of the target languages explored) have the highest relative gains, turning around a very poor baseline across all metrics. The COMET score for KO increased to 84.3 compared to the average of 84.5 in the 100k+ datasets for PR-BR, DE, FI, and CS, resulting in KO’s comparable performance to the high resource languages, i.e. PT-BR and DE. 

These results probably relate not only to the resource level of the language but also to the amount of Korean data in the Llama 3 training recipe. According to MetaAI, “over 5\% of the Llama 3 pre-training dataset consists of high-quality non-English data that covers over 30 languages” \citep{noauthor_meta_2024}. While the Llama Team provides more detail on the training and data mix Llama 3, the exact proportion of Korean data is not discussed \citep{llama_team_llama_2024}. Our baseline metrics suggest that Korean does not feature highly on that list given that it scores significantly lower than all other languages. This might be attributed to the fact that there were not enough examples to produce a firm understanding of the language but enough to provide a foundation that heavily benefited from fine-tuning. As mentioned, this is an assumption as we lack sufficiently detailed information on the training recipe.

When looking at the target languages, we note that PT-BR shows the best performance at 14.7k and 100k+ dataset. This indicates that, even for a well-resourced language, the foundation model gained a strong understanding of the language during pre-training. However, it did not seem to benefit as much from fine-tuning as KO, a lower resource language. This corroborates the finding that resource level is a strong determiner of LLM MT performance \citep{robinson-etal-2023-chatgpt}.

\subsection{Human Evaluation}


Regarding the human evaluation, the qualitative comments from the translators reveal that the largest model struggles with ambiguity. Evaluators mention that segments that lacked complete information needed to be completely reworked. For example, the segment, ``Get basic, step-by-step instructions to learn" lacks a final object, which impacts the translation. While human translators often face and resolve such ambiguities through research or decision-making with incomplete information, the model processes segments in isolation, unable to access potentially clarifying context from adjacent segments. This limitation provides insight into the model's performance in real-world translation scenarios.

\section{Conclusions}

Fine-tuning on TMs has been demonstrated to enhance the performance of LLMs in MT tasks. In this paper, we investigate the relationship between automatic metric results and training set sizes to identify the optimal balance where resource investment yields the most significant improvements in translation quality. In our experiments, it has become evident that fine-tuning on training datasets whose size is larger than 5k examples returned increasingly better results in 19 out of the 20 language-training set size combinations explored.

By leveraging TMs, the model becomes more adept at recognising and reproducing previously translated segments, their style, and terminology. Furthermore, fine-tuning on TM data helps the model adapt to specialised fields.

The test and training sets used come from a much narrower corpus of data than in similar experiments that deal with wider domains, i.e. medicine \citep{moslem_fine-tuning_2023}. The hyper-specific nature of the training data employed in our approach may partly explain the promising results. We therefore leverage the advantage that smaller models licensed for business-use offer; they can be adapted several times over for narrow and specific domains, as well as multiple languages with little investment, instead of aiming for a more general purpose or multilingual model. The hyper-specific purpose of our trained model, i.e. one language direction and a narrow domain, suits the size and easiness of training of an 8B parameter mode.  

Being a commonly experienced scenario in the localization industry, this is an under-explored approach that organisations could be pursuing in order to make the most out of their access to TMs and LLMs for MT in order to obtain the best possible return on investment when leveraging their previously human-translated material.

Low-resource languages seem to be in a perfect position to benefit from leveraging small business-friendly models, like Llama 3 8B. The gains in automatic metric results for KO are substantially higher for high resource languages, like PT-BR and DE, returning the highest increase in performance compared to the metrics obtained from training on similar set sizes in those languages. KO observes an increase of 130\% on COMET from the baseline to the 100k+ dataset, whereas the average increase amongst the other target languages is 46\% (cf. Table \ref{tab:results}).

It is important to mention that, just as \cite{wu_evaluating_2024} acknowledges the FLORES-200 dataset leakage into Llama 3, it is possible that some of our test set was also scraped by the Llama 3 models, as parts of the material were published online prior to the Llama 3 family’s pre-training. We face the same challenge as the whole AI researching community, forced to either constantly come up with new test sets or simply acknowledge the potential leakage of test data \citep{xu2024benchmarking}. We urge large tech companies to disclose at a minimum the test sets that were not ingested by their models for the benefit of the whole community. We acknowledge the Llama Team's leadership in this area \citep{llama_team_llama_2024}.

\section{Future Work}



Future work in the area may benefit from the introduction of checkpoints during training and subsequent intermediate evaluation would enable the visualisation of a clearer learning curve, and the identification of potential dips in performance and points of diminishing returns. This approach would facilitate the analysis and allow for a finer and more efficient evaluation process.

In the future, we aim to obtain a bespoke test set directly from the organisation that owns the TMs. This tailored test set would consist of examples specifically designed in-house according to strict guidelines, ensuring they are completely original and reflective of the organisation's unique requirements and style. By using a bespoke and unseen test set, we can more accurately assess the performance of our fine-tuned models in a real-world context.

Finally, further investigation is required with regard to the training hyperparameters across the different dataset sizes in order to obtain better results with smaller training sets under 5k examples. Several strategies can be explored to optimise performance on smaller datasets. Adjustments such as modifying the dropout rates to prevent overfitting, applying regularisation techniques to enhance model generalisation, and fine-tuning the learning rate to ensure efficient convergence can be particularly beneficial in this case.

\section{Acknowledgements}

This research is supported by Science Foundation Ireland through ADAPT Centre (Grant No. 13/RC/2106) (\url{www.adaptcentre.ie}) at Dublin City University. We thank Alpha-CRC for their essential collaboration.

\begin{small}
\bibliographystyle{apalike}
\bibliography{amta2024,anthology,paperpile}
\end{small}
\end{multicols}

\newpage
\appendix

\section{Appendix A}
\label{sec:appendix-a}

\begin{table*}[htb]
\centering
\begin{tabular}{lllll}
\toprule
\textbf{Lang} & \textbf{Size} & \textbf{Loss} & \textbf{Dev Loss} & \textbf{Train Runtime}
 \\
\midrule
 & 1k & 1.4922 & 1.0706 & 69 \\
 & 2k & 0.8271 & 0.8290 & 120 \\
\textbf{PT-BR} & 5k & 0.7506 & 0.6325 & 290 \\
 & 10k & 0.4222 & 0.5337 & 551 \\
 & 14.7k & 0.4918 & 0.4714 & 820 \\
 & 100k+ & 0.6031 & 0.5964 & 8423 \\
\midrule
 & 1k & 1.5428 & 1.0795 & 70 \\
 & 2k & 0.9446 & 0.8880 & 124 \\
\textbf{CS} & 5k & 0.6643 & 0.6586 & 296 \\
 & 10k & 0.6475 & 0.5880 & 574 \\
 & 14.7k & 0.5346 & 0.5169 & 837 \\
 & 100k+ & 0.5600 & 0.4800 & 8000* \\
\midrule
 & 1k & 1.5342 & 1.1519 & 71 \\
 & 2k & 0.9631 & 0.9602 & 125 \\
\textbf{FI} & 5k & 0.5876 & 0.6286 & 302 \\
 & 10k & 0.5662 & 0.5874 & 589 \\
 & 14.7k & 0.3996 & 0.5138 & 866 \\
 & 100k+ & 0.5964 & 0.5867 & 8241 \\
\midrule
 & 1k & 1.5551 & 1.1397 & 69 \\
 & 2k & 0.9591 & 0.9301 & 121 \\
\textbf{DE} & 5k & 0.4371 & 0.6426 & 290 \\
 & 10k & 0.4553 & 0.5639 & 550 \\
 & 14.7k & 0.5310 & 0.5037 & 819 \\
 & 100k+ & 0.6672 & 0.6603 & 8000* \\
\midrule
 & 1k & 1.5851 & 1.0651 & 67 \\
 & 2k & 0.7765 & 0.7733 & 120 \\
\textbf{KO} & 5k & 0.6086 & 0.6340 & 270 \\
 & 10k & 0.4662 & 0.5666 & 543 \\
 & 14.7k & 0.4167 & 0.4923 & 807 \\
 & 100k+ & 0.7822 & 0.7052 & 5791 \\
\bottomrule
\end{tabular}
\caption{Training Details by Language. Train Runtime is measured in seconds. Starred numbers are estimates.}
\label{tab:model-performance}
\end{table*}

\section{Appendix B}
\label{sec:appendix-b}

\subsection{Special Token Descriptions}
\begin{quote}
\begin{footnotesize}
\begin{it}
$<|$begin\_of\_text$|>:$ This is equivalent to the BOS token.\\
$<|$eot\_id$|>:$ This signifies the end of the message in a turn.\\
$<|$start\_header\_id$|>$\{role\}$<|$end\_header\_id$|>:$ These tokens enclose the role for a particular message. The possible roles can be: system, user, assistant. \\
$<|$end\_of\_text$|>:$ This is equivalent to the EOS token.
\end{it}
\end{footnotesize}
\end{quote}

\subsection{Prompt}

\begin{quote}
\begin{footnotesize}
\begin{it}
$<|$begin\_of\_text$|>$
$<|$start\_header\_id$|>$system$<|$end\_header\_id$|>$\\ \\
You are a helpful AI assistant for translation from \{source\_language\} to \{target\_language\}. You MUST answer with the following JSON scheme: \{``translation'': ``string''\}
$<|$eot\_id$|>$\\
$<|$start\_header\_id$|>$user$<|$end\_header\_id$|>$ \\ \\
\{source\_sentence\}$<|$eot\_id$|>$$<|$start\_header\_id$|>$assistant$<|$end\_header\_id$|>$
\end{it}
\end{footnotesize}
\end{quote}

\subsection{Training Prompt}

\begin{quote}
\begin{footnotesize}
\begin{it}
$<|$begin\_of\_text$|>$$<|$start\_header\_id$|>$system$<|$end\_header\_id$|>$ \\

You are a helpful AI assistant for translation from \{source\_language\} to \{target\_language\}. You MUST answer with the following JSON scheme: \{``translation'': ``string''\} $<|$eot\_id$|>$ 

$<|$start\_header\_id$|>$user$<|$end\_header\_id$|>$
\{source\_sentence\}$<|$eot\_id$|>$ \\

$<|$start\_header\_id$|>$assistant$<|$end\_header\_id$|>$\textbf{\{target\_sentence\}$<|$end\_of\_text$|>$}
\end{it}
\end{footnotesize}
\end{quote}

\section{Appendix C}
\label{sec:appendix-c}

\begin{table*}[!ht]
\centering
\begin{tabular*}{\textwidth}{@{\extracolsep{\fill}} ll}
\toprule
{\textbf{Inference Parameters}} \\
\midrule
sampling\_topk                        & 1 \\
max\_batch\_size                      & 8096 \\
min\_length                           & 1 \\
max\_length                           & double the source length \\
\bottomrule
\end{tabular*}
\caption{CTranslate2 Inference Parameters.}
\label{tab:inferenceparameters}
\end{table*}

\newpage
\section{Appendix D}
\label{sec:appendix-d}

\begin{table*}[!ht]
\centering
\begin{small}
\begin{tabular}{l*{2}{>{\raggedleft\arraybackslash}p{5em}}}
\textbf{Lang} & \textbf{Data Size} & \textbf{Comet-Kiwi $\uparrow$} \\
\toprule
& 1k & 77.95 \\
& 2k & 77.81 \\
\textbf{PT-BR} & 5k & 77.65 \\
& 10k & 77.77 \\
& 14.7k & 78.92 \\
\midrule
& 1k & 79.71 \\
& 2k & 78.98 \\
\textbf{CS} & 5k & 78.57 \\
& 10k & 78.78 \\
& 14.7k & 79.71 \\
\midrule
& 1k & 78.58 \\
& 2k & 78.22 \\
\textbf{DE} & 5k & 78.34 \\
& 10k & 78.21 \\
& 14.7k & 78.73 \\
\midrule
& 1k & 79.58 \\
& 2k & 78.70 \\
\textbf{FI} & 5k & 78.12 \\
& 10k & 78.54 \\
& 14.7k & 79.58 \\
\midrule
& 1k & 81.93 \\
& 2k & 81.56 \\
\textbf{KO} & 5k & 81.20 \\
& 10k & 81.22 \\
& 14.7k & 81.55 \\
\bottomrule
\end{tabular}
\end{small}
\caption{Quality Evaluation results of training datasets of different sizes using Comet-Kiwi metric.}
\label{tab:QEresults}
\end{table*}

\end{document}